\documentclass[journal]{IEEEtran}

\usepackage[noadjust]{cite}

\ifCLASSINFOpdf
   \usepackage[pdftex]{graphicx}
   \graphicspath{{Figs/}}
   \DeclareGraphicsExtensions{.pdf}
\else
\fi
\usepackage{amsmath}
\usepackage{bm,amssymb}

\usepackage{algorithm}
\usepackage{algorithmic}

\usepackage{array}

\usepackage{makecell}

\usepackage{fixltx2e}

\usepackage{placeins} 

\usepackage{stfloats}

\newcommand{\etal}{\emph{et al.}}
\newcommand{\eg}{e.g.}
\newcommand{\ie}{i.e.}

\begin{document}

\title{Discriminative Sparse Neighbor Approximation \\ for Imbalanced Learning}

\author{Chen~Huang,~Chen~Change~Loy,~\IEEEmembership{Member,~IEEE,}~and~Xiaoou~Tang,~\IEEEmembership{Fellow,~IEEE}
\thanks{All the authors are with the Department of Information Engineering, The Chinese University of Hong Kong, Hong Kong.~E-mail: \{chuang,ccloy,xtang\}@ie.cuhk.edu.hk.}}

\maketitle

\begin{abstract}
Data imbalance is common in many vision tasks where one or more classes are rare. Without addressing this issue conventional methods tend to be biased toward the majority class with poor predictive accuracy for the minority class. These methods further deteriorate on small, imbalanced data that has a large degree of class overlap.
In this study, we propose a novel discriminative sparse neighbor approximation (DSNA) method to ameliorate the effect of class-imbalance during prediction.
Specifically, given a test sample, we first traverse it through a cost-sensitive decision forest to collect a good subset of training examples in its local neighborhood. Then we generate from this subset several class-discriminating but overlapping clusters and model each as an affine subspace.
From these subspaces, the proposed DSNA iteratively seeks an optimal approximation of the test sample and outputs an unbiased prediction.
We show that our method not only effectively mitigates the imbalance issue, but also allows the prediction to extrapolate to unseen data. The latter capability is crucial for achieving accurate prediction on small dataset with limited samples.
%
%
The proposed imbalanced learning method can be applied to both classification and regression tasks at a wide range of imbalance levels. It significantly outperforms the state-of-the-art methods that do not possess an imbalance handling mechanism, and is found to perform comparably or even better than recent deep learning methods by using hand-crafted features only.

\end{abstract}


%

\section{Introduction}
\label{sec1}

\begin{figure}[t]
\centering
\includegraphics[width=1.0 \linewidth]{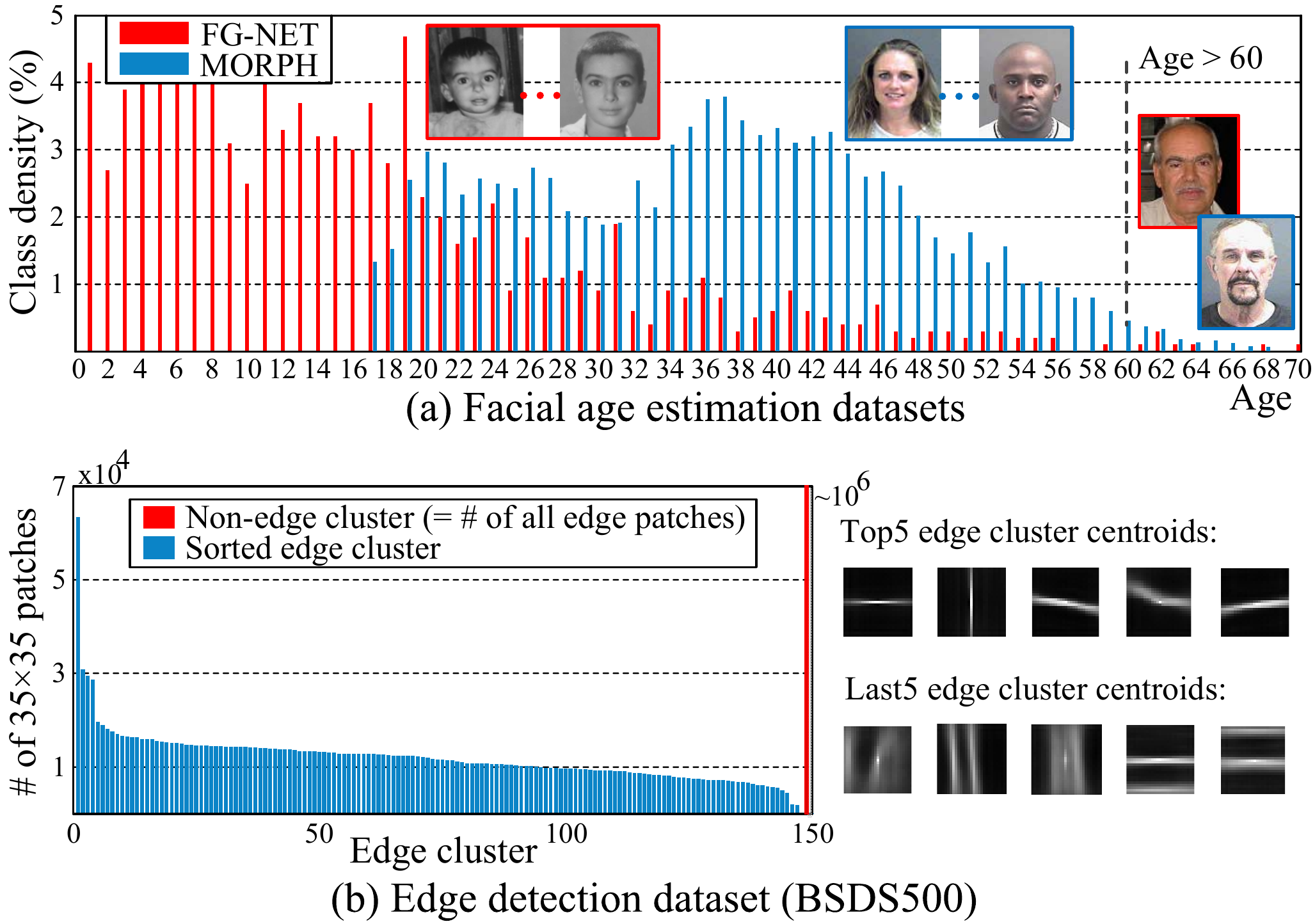}
\caption{Data imbalance in (a) age estimation and (b) edge detection. The given datasets characterize the underlying imbalanced distributions that can be seen as intrinsic in these problems.}
\label{fig1}
\end{figure}

\IEEEPARstart{D}{ata} imbalance exists in many vision tasks ranging from low-level edge detection~\cite{Pablo11} to high-level facial age estimation~\cite{Geng07} and head pose estimation~\cite{AghajanianP09}.
For instance, in age estimation, there are often many more images of the youth than the old on the widely used FG-NET~\cite{Geng07} and MORPH~\cite{Chang11} datasets.
In edge detection, various image edge structures~\cite{LimCVPR13} obey a power-law distribution, as shown in Figure~\ref{fig1}.
Without handling this imbalance issue conventional vision algorithms have a strong learning bias towards the majority class with poor predictive accuracy for the minority class, usually of equal or more interest (\eg~rare edges may convey the most important semantic information about natural images).

\begin{figure}[t]
\centering
\includegraphics[width=1.0 \linewidth]{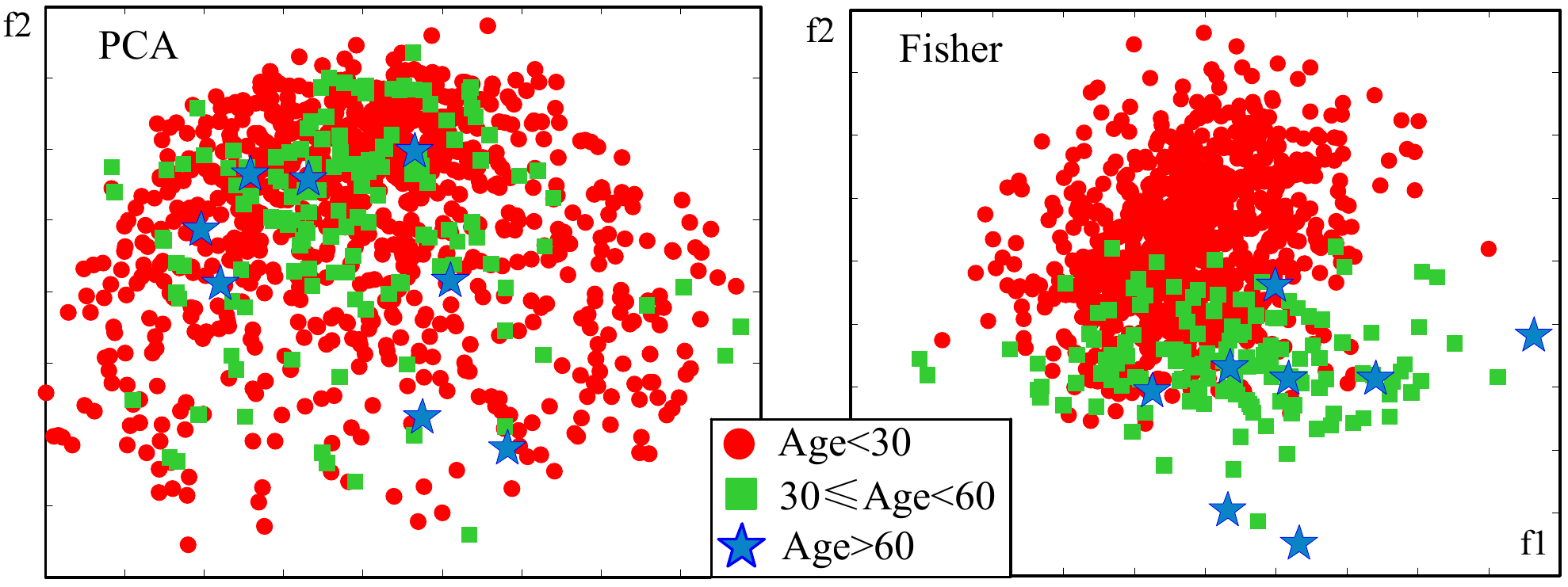}
\caption{2D distributions of age data after PCA and LDA on FG-NET to highlight the class overlap issue.}
\label{fig2}
\end{figure}

The insufficient learning for the minority class is due to the complete lack of representation by a limited number of or even no examples, especially in the presence of small datasets. For instance, FG-NET age dataset has 1002 images in total
with only 8 images over $60$ years old. Certain age classes of $60+$ ages have no images at all. This reveals a bigger challenge on unseen data extrapolation from the few minority class samples that usually have high variability. Even worse, the small imbalanced datasets can be accompanied by the class overlap problem.
We plot the PCA and Fisher embeddings of FG-NET in Figure~\ref{fig2} to illustrate this problem.
From the figure, it is evident that training a robust classifier or regressor capable of handling old ages is indeed a hard problem: (i) the corresponding minority class (blue star) contains insufficient samples for learning, (ii) these samples have high degree of variability which is hard to model, (iii) there is a severe class overlap between the rare samples and those from majority classes, further compounding the learning difficulty.
Consequently, if we look into the local neighborhood of a minority class sample, it is very likely to be dominated by the majority class samples. Its weak local boundary would bias the prediction towards the majority class.

There are three common approaches to counter the negative impact of data imbalance: resampling~\cite{Chawla02,He09}, cost-sensitive learning~\cite{Drummond03,LiL06,Zadrozny03} and ensemble learning~\cite{Chen04,Ting00}.
Resampling approaches aim to make class priors equal by under-sampling the majority class or over-sampling the minority class (or both~\cite{Chawla02}).
These methods can easily eliminate valuable information or introduce noise respectively.
Cost-sensitive learning is often reported to outperform random re-sampling by adjusting misclassification costs, however the true costs are often unknown. An effective technique for further improvement is to resort to ensemble learning~\cite{Breiman01}. Chen \etal~\cite{Chen04} combined bagging and weighted decision trees to generate a re-weighted version of random forest. We show in our experiments that the aforementioned strategies fall short in handling complex imbalanced data.
Beyond empirical performance, the above approaches have two common drawbacks: 1) They are designed for either classification~\cite{Chawla02,Chen04,Drummond03,He09,Ting00,Zadrozny03} or regression~\cite{LiL06} without a universal solution to both. 2) They have a limited ability to account for unseen appearances or extrapolate novel labels on the observed space. This is critical in the typical case of small imbalanced datasets where the minority class is under-represented by an excessively reduced number of or even no samples/labels.

In this paper we address the problems of data imbalance {\em and} unseen data extrapolation using a data-driven approach. The approach can be applied to {\em both} classification and regression scenarios.
The key idea of our approach is intuitive -- given a test sample, we first locate for it a `safe' local neighborhood. This local neighborhood is formed by training samples, which are carefully mined so as to provide a relatively large coverage of minority class samples compared to the full training space. But overall, this space is tight and is less probable to be invaded by imposter samples\footnote{An imposter sample is defined as the one from a different class w.r.t. the test sample}. We show that this `safe' local neighborhood can be constructed via a cost-sensitive decision forest.
However, the local neighborhood may still be overwhelmed by majority classes especially when the minority ones are absolutely rare. Thus prediction by simple voting or averaging within it could easily smooth out the minority class samples.
To this end, we further partition the local neighborhood into several discriminative but soft clusters with overlaps permitted. This process provides purer clusters eliminating the undesired class domination.

\begin{figure}[t]
\begin{center}
\includegraphics[width=0.9 \linewidth]{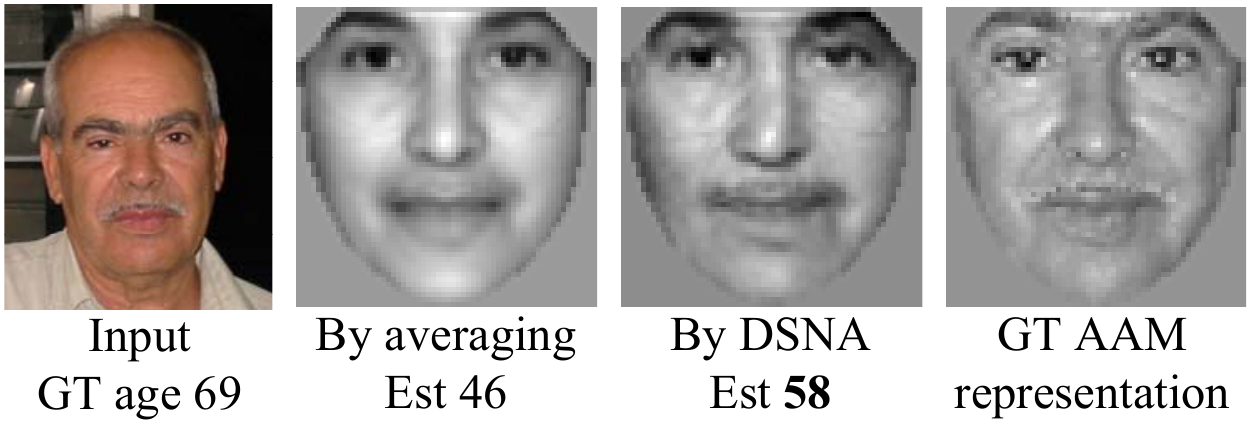}
\end{center}
\vskip -0.35cm
\caption{A visualization of age estimation result when neither the testing appearance nor age label is observed during training. Averaging provides a crude way of estimating the face appearance (AAM, Active Appearance Model) from the nearest neighbors in the training set. The proposed discriminative sparse neighbor approximation (DSNA) provides a more robust estimation, thus an age value closer to the Ground Truth (GT).}
\label{fig3}
\end{figure}

Subsequently, we propose a new Discriminative Sparse Neighbor Approximation (DSNA) method that allows robust prediction from our formed clusters.
The clusters are all modelled as affine subspaces to account for unseen appearances in a similar spirit of~\cite{Hu11}.
The core of DSNA is a new cost function and a joint optimization approach to iteratively determine the best affine subspace that best approximates the test sample with the help of associated sparse neighbors. From the found neighbors and their approximating coefficients, we can transfer and combine their labels to achieve a robust prediction despite class-imbalanced issue. %
Figure~\ref{fig3} illustrates the effectiveness of DSNA in an age estimation example.

In summary, the main contributions of this paper are:
\begin{itemize}
  \item A new discriminative sparse neighbor approximation (DSNA) method is proposed for unbiased predictions with preserved discriminative and extrapolative ability given class-imbalanced data.
   \item To facilitate robust predictions via DSNA, we formulate an effective way of constructing a safe local neighorhood through a cost-sensitive decision forest framework.
  \item The proposed method is applied to the vision tasks of age estimation (regression), head pose estimation (regression) and edge detection (classification) with varying degree of data imbalance and amount of data. It advances the state-of-the-art, sometimes considerably, across all tasks especially on highly imbalanced ones. It comes at only modest extra computational burden, showing its potential as a fast and general framework for imbalanced learning. Our results are particularly impressive when favorably compared to recent deep learning methods \cite{KongJY14,Dong14,KivinenWH14,Ganin14,gberta15_CVPR,Shen15,gberta15_ICCV,xie15} as our method is built with no deeply learned features, but with the imbalance handling mechanisms absent in these deep models.
\end{itemize}

The rest of the paper is organized as follows. Section \ref{sec2} briefly reviews related work on imbalanced learning and the considered vision tasks. Section \ref{sec3} details the major components of the proposed framework of imbalanced learning. Section \ref{sec4} presents the results in ablation tests and three imbalanced vision tasks to highlight the benefit of each proposed component and their advantages over competing methods in different tasks. Section \ref{sec5} concludes the paper.

\section{Related Work}
\label{sec2}
Much effort for imbalanced learning in the machine learning community has been devoted to resampling approaches~\cite{He09} that randomly under-sample the majority class or over-sample the minority. Other smart resampling techniques are also available (please refer to~\cite{He09} for a comprehensive survey). Generally under-sampling may remove valuable information and over-sampling easily introduces noise with overfitting risks. Additionally, random over-sampling does not increase information by only replication, so it does not solve the fundamental ``lack of data'' issue. SMOTE~\cite{Chawla02}, on the other hand, creates new examples by interpolating neighboring minority class instances. However, it is error-prone to interpolate noisy or borderline examples. Therefore under-sampling is often preferred to over-sampling~\cite{Drummond03}, but is not suitable for small datasets (\eg~FG-NET) due to its caused information loss.

Cost-sensitive learning~\cite{Drummond03,LiL06,Zadrozny03} as an alternative is closely related to resampling. Instead of manipulating samples at the data level, it adjusts misclassification costs at the algorithmic level and imposes heavier penalty on misclassifying the minority class. For example, Li and Lin~\cite{LiL06} proposed RED-SVM to use the label-sensitive costs in the ordinal regression problem. In~\cite{Zhang14}, a scaling kernel with the standard SVM is used to improve the classification on imbalanced datasets. Zadrozny \etal~\cite{Zadrozny03} combined cost sensitivity with ensemble approaches to further improve classification accuracy. Chen \etal~\cite{Chen04} formed an ensemble of cost-sensitive decision trees by weighting the Gini criterion during the node splitting as well as final tree aggregation. We similarly grow cost-sensitive random trees but more generalized and principled ones, and propose a more discriminative and extrapolative ``aggregation'' scheme that proves necessary for complex imbalanced data.

The methods of~\cite{Zadrozny03,Chen04} already show the effectiveness of using classifier ensemble in the context of imbalanced data~\cite{Chen04,Ting00}. Bagging and Boosting are the most popular ensemble strategies~\cite{Breiman01}. Boosting (\eg~\cite{Ting00,Li14}) can easily embed the cost sensitivities in example weights according to the misclassification costs. Li \etal~\cite{Li14} further combined boosting with the training of an extreme learning machine. Generally boosting is vulnerable to noise and more prone to overfitting, which can be better addressed by Bagging~\cite{Breiman01}. Our method based on the improved random forest is essentially a Bagging-based method, thus shares this advantage.

Although much progress has been made on the vision tasks such as age estimation, head pose estimation and edge detection, relatively few works have studied the direct impacts of data imbalance on these tasks. Next we briefly discuss their representative works.

\noindent
{\bf Age estimation:} There are three main groups of age estimation methods: classification~\cite{Geng07,Ren14,Choi11}, regression~\cite{Guo11,Guo13,LiL06,Zhang10,Chao13}, and ranking~\cite{Chang11,Yan07,Chang15} methods. OHRank~\cite{Chang11,Chang15} surpasses previous classification- and regression-based methods by utilizing ordering information and cost sensitivities. However, the imbalance issue is neglected especially when designing the ordered binary classifiers at the youngest and oldest ages. Some recent state-of-the-arts focus on advanced feature extraction~\cite{Guo09,Ren14,Chang15}, including applying convolutional neural network (CNN)~\cite{Dong14,KongJY14} to automatically learn deep features instead of using hand-crafted ones. Unfortunately strong biases are still observed on imbalanced datasets, and we provide here an explicit solution to imbalanced learning with better results, using no deep features. Only three papers~\cite{Ke13,Geng13,Chao13}, as far as we know, consider data imbalance and sparseness when estimating ages. lsRCA~\cite{Chao13} simply balances the number of used neighbors from each class to compute the similarity matrix for LPP (Locality Preserving Projection)-based dimensionality reduction. In~\cite{Ke13,Geng13}, the imbalance is only mitigated by leveraging adjacent labels in implicit ways, respectively via modeling cumulative attribute space and label distribution. We will show the advantage of our explicit solution and that of our built-in extrapolative mechanism for possible missing data/labels.

\noindent
{\bf Head pose estimation:} Methods for head pose estimation from 2D images can be categorized into two main groups: classification~\cite{Huang10} and regression~\cite{drouard15,Fenzi13,Haj12,Hara14,Liu14}, with regression being more attractive for its continuous output. We refer readers to~\cite{Murphy09} for a comprehensive survey. Random forest is a popular choice for pose estimation in both classification~\cite{Huang10} and regression~\cite{Hara14} settings. It is also applied to depth images~\cite{Fanelli13}. To our knowledge, the inherent imbalance in pose data~\cite{AghajanianP09} is seldom addressed again. Note on many pose datasets such as Pointing'04, the sparse data sampling (with typical pose intervals of $10^{\circ}+$) makes learning even more difficult.

\noindent
{\bf Edge detection:} State-of-the-art edge detection methods \cite{Pablo11,APBMM2014,Dollar15,Isola14,LimCVPR13,ren12,ren13,HallmanF15} mostly use engineered gradient features to classify edge pixels/patches. Recent CNN-based methods \cite{Ganin14,KivinenWH14,Shen15,gberta15_CVPR,gberta15_ICCV,xie15} achieve top results by learning deep features. Due to the large variety of edge structures, it is usually very hard to learn an ideal binary classifier to separate edges as one class from the non-edge class. Therefore some methods first cluster edge patches into compact subclasses (\eg~\cite{LimCVPR13,Shen15}), and cast edge detection as a multi-way classification problem (\ie~to predict whether an input patch belongs to each edge subclass or the non-edge class) so as to implicitly solve the binary task. For the same reason, the numbers of ``positive'' and ``negative'' patches are commonly set to be equal to facilitate the binary goal. However, this results in a severe imbalance between each edge subclass and the dominant negative one (see Figure~\ref{fig1}(b)), which is barely addressed properly by the above methods. Consequently, biased predictions tend to occur with low edge recall or damage of fine edge structures in those rare subclasses.

Another limitation of existing methods is that they cannot well predict the unseen edge structures from a novel class. For example, Sketch Tokens~\cite{LimCVPR13} only predict from a pre-defined set of edge classes based on random forest. Structured Edge (SE) detector~\cite{Dollar15} can model more subtle edge variations in a structured forest framework without the finite-class assumption, but still can only infer the edge structures observed during training. Although this problem is ameliorated by merging predicted structures while testing, it is in sharp contrast to our explicit DSNA method that empowers random forests to extrapolate.

\begin{figure*}
\begin{center}
\includegraphics[width=1.0 \linewidth]{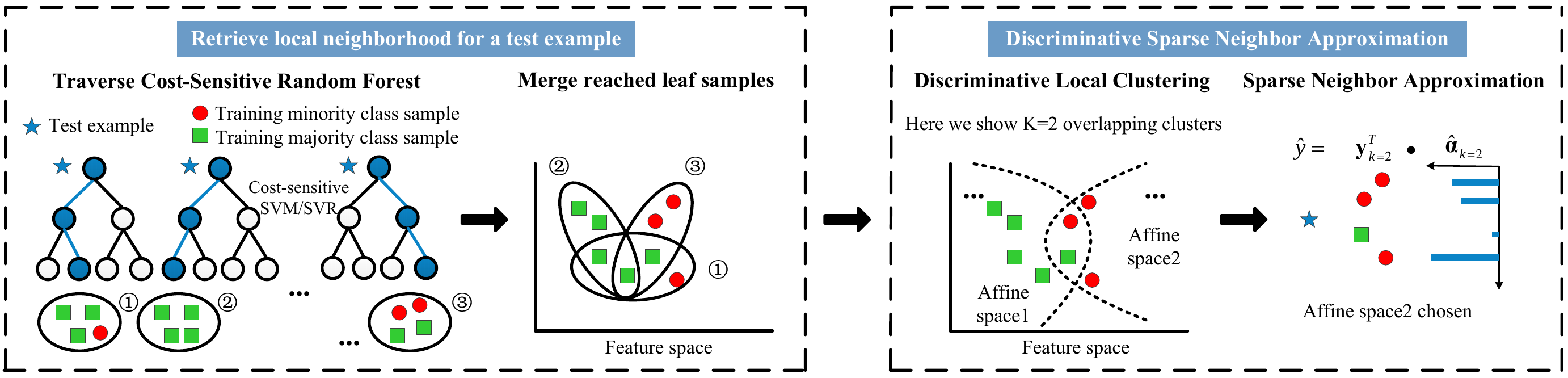}
\end{center}
\vskip -0.25cm
\caption{The overall pipeline of our CS-RF-induced DSNA method.}
\label{fig4}
\end{figure*}

\section{Methodology}
\label{sec3}

The proposed discriminative sparse neighbor approximation (DSNA) aims to provide unbiased predictions given a class-imbalanced dataset.
More precisely, given a training set $\mathcal{D} = \left\{s_i=(\bm{x}_i,y_i)\right\}_{i=1}^N$, where $\bm{x}_i\in\mathbb{R}^{D}$ is the feature vector of sample $s_i$ and $y_i$ the label, our problem can be formulated as learning a function $F(\bm{x})\rightarrow y$ to make unbiased predictions even in the presence of severely imbalanced and small datasets. The label $y\in\mathcal{C}$ refers to the class index (\eg~edge class) for classification or a numeric value (\eg~age and pose angle) for regression.

For a query $\mathbf{q}$, the key steps of DSNA are to draw a well localized neighborhood with a selective set of training data that is less probable to be invaded by imposter samples (\ie~from a different class w.r.t. the test sample), then follow the ``divide and conquer'' idea to perform a class-discriminative local clustering to obtain much purer clusters (modeled as affine subspaces) without undesired class domination, and finally choose the best cluster and use its member labels to predict.

The overall pipeline is shown in Figure~\ref{fig4}. The pipeline begins with a cost-sensitive random decision Forest (CS-RF), which takes care of generating an initial good local neighborhood at leaf nodes, in order to reduce unnecessary distractions from majority classes. We retrieve all the leaf samples for a test instance, aiming to gain as more coverage of relevant minority samples as possible.
The DSNA component starts with discriminative local clustering, and performs a sparse approximation to iteratively output unbiased predictions. To enable extrapolated prediction for unseen appearances, we model the found clusters as affine subspaces and extrapolate a prediction from them.

In the following, we first present the DSNA approach which is the key of this paper. We make the assumption that local neighborhood of training data is already available. We then describe the use of cost-sensitive random decision forest to obtain the local neighborhood.


\subsection{Discriminative Sparse Neighbor Approximation}
\label{subsec:dsna}

\noindent
\textbf{Discriminative local clustering} -
The first step of DSNA is to perform discriminative clustering within the local data neighborhood of a test sample.
Suppose we have an initially retrieved local data neighborhood at hand, which can be noisy and class-imbalanced. This local neighborhood can be represented as
\begin{equation}
\label{eqn:local_neighborhood}
\mathcal{R} = \left\{ s_i\right\}_{i=1}^{M},\;\;\; \mathcal{R} \subset  \mathcal{D}, \;\; M < N.
\end{equation}
Intuitively, the samples in $\mathcal{R}$ are close to the test sample based on some notions of metric or non-metric distance.
Our objective is to separate the samples in $\mathcal{R}$ based on their different class labels so as to pave the way for unbiased prediction of the test sample.
We shall choose a clustering technique that possesses two desirable properties to achieve this goal: 1) It should generate discriminative clusters from one of which unbiased predictions can be made. 2) The found clusters should have adequate descriptiveness to account for unseen data patterns.

We achieve the aforementioned goal through a simple yet effective extension of K-means.
It differs from the standard K-means in two respects. First, the inter-point distance $\widetilde{d}\left( \bm{x}_i,\bm{x}_j \right)$ between $\bm{x}_i$ and $\bm{x}_j$ is label-aware:
\begin{equation}
\label{eq1}
\widetilde{d}\left( \bm{x}_i,\bm{x}_j \right)= \left\{ d\left( \bm{x}_i,\bm{x}_j \right)*\bm{1}\left(y_i \neq y_j\right) \; \mathrm{for\ classification}, \atop d\left( \bm{x}_i,\bm{x}_j \right)*g\left(|y_i-y_j|\right) \;\quad \mathrm{for\ regression}, \right.
\end{equation}
where $d(\cdot,\cdot)$ is the Euclidean distance, $\bm{1}(\cdot)$ is an indicator function, $g(y)=\tau y/(max\{y\}-y+\mathrm{eps})$ is a reciprocal increasing function with $\tau$ the trade-off parameter, and $\mathrm{eps}$ a small positive number to prevent overflow.
The label-aware distance makes clustering discriminative by preferring the ``same-class'' data-pairs over those from different classes. In the extreme case, under classification scenarios for example, it forms clusters $\{\mathcal{L}_k\}_{k=1}^K$ each purely from one class even when the cluster members differ remarkably in appearances, which is suitable for classification.

Considering it is highly possible that the ``pure'' clusters in small imbalanced problems have limited samples, especially those mostly with the minority class samples, such clustering is actually not desirable for data/label extrapolation purposes. Hence, we allow cluster overlap by relaxing the cluster assignment of sample $\bm{x}_i$. Instead of assigning it solely to the nearest cluster centroid, we choose more than one centroids with distances slightly larger than
%
%
the minimum distance in each K-means optimization iteration. This results in overlapping clusters each containing some ``inter-class'' samples. Such samples have complementary appearances to those ``same-class'' ones for enriching cluster representations.


\vspace{0.1cm}
\noindent \textbf{Sparse neighbor approximation} -
The previous step generates $K$ overlapping clusters $\{\mathcal{L}_k\}_{k=1}^K$ with their feature matrices $\{\bm{L}_k\}_{k=1}^K$ and labels $\{\bm{y}_k\}_{k=1}^K$. Our problem becomes how to discriminatively predict the label of a query $\bm{q}$ and extrapolate to its possibly unseen appearance simultaneously.

To address this problem, we model each cluster by an affine hull model $\mathcal{AH}_k$~\cite{Hu11} that is able to account for unseen data of different modes, and then choose the best prediction returned by them. Every single $\mathcal{AH}_k$ covers all possible affine combinations of its belonging samples and can be parameterized as:
\begin{equation}
\label{eq2}
\mathcal{AH}_k = \left\{ \bm{x} = \bm{\mu}_k+\bm{U}_k\bm{v}_k,k=1,\ldots,K \right\},
\end{equation}
where $\bm{\mu}_k=\sum_{\bm{x}_i\in\mathcal{L}_k}\bm{x}_i/|\mathcal{L}_k|$ is the centroid, $\bm{U}_k$ is the orthonormal basis obtained from the SVD of centered $\bm{L}_k$, and $\bm{v}_k$ is the coefficient vector.

Note that to predict the class label of query $\mathbf{q}$, we need to know which cluster the query should be assigned to. Thus Eq.~\ref{eq2} only provides a partial answer to our problem, since the cluster index $k$ remains unknown.
To this end, we formulate a joint optimization problem for simultaneuously finding the belonging cluster of the query and its affine hull approximation:
\begin{eqnarray}
\label{eq5}
&&\!\!\!\!\!\!\!\!\!\!\!\!\!\!\underset{k,\bm{v}_k,\bm{\alpha}_k}{\min} \left\| \bm{\mu}_k+\bm{U}_k\bm{v}_k- \bm{L}_k \bm{\alpha}_k\right\|_2 + \lambda\left\|\bm{\alpha}_k\right\|_1 + \gamma\left\|\bm{\alpha}_k-\overline{\bm{\alpha}}_k\right\|_1,  \nonumber\\
&&\!\!\!\!\!\!s.t. \; \left\| \bm{q}-(\bm{\mu}_k+\bm{U}_k\bm{v}_k) \right\|_2 \leq \varepsilon,
\end{eqnarray}
where $\varepsilon\geq0$, and $\lambda$ and $\gamma$ are regularization parameters.


We explain the objective function as follows:

\textit{First term} - This term approximates $\bm{q}$ over the $k^{th}$ cluster using the cluster's affine subspace as well as the feature matrix of associated member samples $\bm{L}_k$.
This term is motivated by affine hull models~\cite{Hu11} but differs significantly in the following aspects:

\noindent i) the affine space is class-aware. In particular, the affine space is learned from our class-discriminating cluster, and we solve for the best approximation among the clusters. A class-aware sparsity constraint is further imposed to promote discrimination (\textit{Third term}).

\noindent ii) the affine space approximation benefits from the enriched descriptiveness of overlapping clusters. 

\vspace{0.1cm}
\textit{Second term} - This term constrains the loose affine approximation by imposing sparsity among the cluster samples. Thus a large drift is avoided when extrapolating $\bm{q}$ on the affine subspace, because we constrain the affine subspace to be near to the observed samples using feature matrix $\bm{L}_k$.
%


\vspace{0.1cm}
\textit{Third term} - This term regularizes the coefficient vector $\bm{\alpha}_k$ so it that  focuses more on the ``same-class'' nearest neighbors, $\mathcal{N}_k=\{ \bm{x}_i\in\mathcal{L}_k:  \widetilde{d}(\bm{x}_i,\bm{q}) \leq \varepsilon_k\}$, which are found by using the class-aware distances in Eq.~\ref{eq1}.
From our experiments, we empirically found that this term is useful to provide stable predictions.
Formally, the $\overline{\bm{\alpha}}_k$ is estimated as:
\begin{equation}
\label{eq4}
\overline{\bm{\alpha}}_k = \sum_{\bm{x}_i\in\mathcal{N}_k} w_i \bm{\alpha}_i, \, w_i \propto \exp(-\widetilde{d}(\bm{x}_i,\bm{q})/h),
\end{equation}
where $h$ is the decay parameter, and $\bm{\alpha}_i$ is the representation coefficient of the $i^{th}$ neighbor with the $i^{th}$ element equal to one and the rest zero.

Eq.~\ref{eq5} can be solved by alternatively seeking the best affine approximation $\min_{k,\bm{v}_k} \left\| \bm{q}-(\bm{\mu}_k+\bm{U}_k\bm{v}_k) \right\|_2$ and the sparse neighbor approximation with two $l_1$-norms:
\begin{equation}
\label{eq6}
\underset{\bm{\alpha}_k}{\min} \left\| \bm{q}- \bm{L}_k \bm{\alpha}_k\right\|_2 + \lambda\left\|\bm{\alpha}_k\right\|_1 + \gamma\left\|\bm{\alpha}_k-\overline{\bm{\alpha}}_k\right\|_1,
\end{equation}
which can be efficiently solved by using the Augmented Lagrange Multiplier (ALM) method~\cite{Bertsekas03}.

With the converged $\hat{\bm{\alpha}}_k$, the label for $\bm{q}$ is finally predicted as $\hat{y}=\bm{y}_k^T\hat{\bm{\alpha}}_k$ for regression or by majority voting for classification (in this case we determine the nonzero entries of thresholded $\hat{\bm{\alpha}}_k$, and vote among the corresponding $\bm{y}_k$). The initial label to start the iterative process is set as the mean or majority vote of $\bm{y}_k$ in the best-fit cluster.


\subsection{Cost-Sensitive Random Decision Forest}
\label{CS-RF}

Returning to the initial step of finding a `safe' local neighborhood, we choose random decision forest for its efficiency and robustness. We first traverse a test example through every trained decision tree and retrieve the respective training samples $\mathcal{R}_t$ stored at the leaf node. Traditional random forest calculates either a class distribution for classification or a local mean for regression from each $\mathcal{R}_t$, and aggregates them as the final prediction. We face two fundamental problems by doing so: in the case of absolute rarity, each $\mathcal{R}_t$ will still predominantly consist of majority classes that make simple aggregation biased to them; or $\mathcal{R}_t$ may form pure but small disjuncts~\cite{He09} of minority class samples leading to overfit.

Therefore, we instead merge all the retrieved leaf sample sets $\{\mathcal{R}_t\}$ into a single one $\mathcal{R}=\cup_t \mathcal{R}_t$, and treat $\mathcal{R}$ as our initial local neighborhood in Eq.~\ref{eqn:local_neighborhood}. Then DSNA is applied to it for prediction as described in Section~\ref{subsec:dsna}. Such a simple merging has the benefit of facilitating our data-driven prediction with as more coverage of relevant minority samples as possible. This can be easily realized thanks to the diversities between merged trees. In fact, random forest has the proved upper bound of generalization error given by~\cite{Breiman01}:
\begin{equation}
\label{eq7}
\epsilon \leq \rho (1-m^2)/m^2,
\end{equation}
where $m$ is the strength of individual trees and $\rho$ is the correlation between decision trees. Hence in order to maintain the low correlation and diversity among trees, we just keep the Bagging nature and feature randomness at internal nodes in the standard random forest.

To make the merged neighborhood less distracted by imposter samples, we focus on improving the strength $m$ of each tree in the context of data imbalance by making the tree cost-sensitive. We have explored different cost-sensitive schemes, such as the re-weighting of nodes as in~\cite{Chen04} and boosting of trees with class costs, but seen marginal effects. We finally came to a modified node splitting rule that can not only take into account the imbalanced distribution, but also can work seamlessly for both classification and regression.

Specifically, we first follow the standard Bagging procedure to grow an ensemble of random trees. Each tree recursively divides the input space into disjoint partitions in a coarse-to-fine manner. The key is to design good splitting functions. For a node $j$ with local samples $\mathcal{S}_j$, a binary function $\phi_j:\mathbb{R}^{D'}\rightarrow \{0,1\}$ is trained on some randomly sampled features ($D'=\sqrt{D}$) and splits into $\mathcal{S}_j^l$ and $\mathcal{S}_j^r$ to maximize the information gain:
\begin{equation}
\label{eq8}
\mathcal{I}(\mathcal{S}_j,\phi_j)=H(\mathcal{S}_j)-\left( \frac{|\mathcal{S}_j^l|}{|\mathcal{S}_j|} H(\mathcal{S}_j^l) + \frac{|\mathcal{S}_j^r|}{|\mathcal{S}_j|} H(\mathcal{S}_j^r) \right),
\end{equation}
where $H(\cdot)$ denotes the class entropy. For regression, information gain can be replaced by the label variance as $H(\mathcal{S})=\sum_y(y-\mu)^2/|\mathcal{S}|$ where $\mu=\sum_y y/|\mathcal{S}|$. Training stops when a maximum depth is reached or if information gain or local sample size $|\mathcal{S}_j|$ falls below a fixed threshold.

The standard node splitting function $\phi_j$ is not necessarily optimal with respect to imbalanced data. To alleviate this problem, in both classification and regression scenarios, we incorporate a cost function $f(\cdot)\geq 0$ into $\phi_j$ that penalizes more heavily on the minority class. We describe in the following the cost function for classification trees and regression trees, respectively.

In classification trees, we first apply the widely used K-means technique~\cite{Dollar15,Hara14} to cluster $\mathcal{S}_j$ into $\{\mathcal{S}_j^k\}_{k=1}^2$, and then the splitting function $\phi_j$ that best preserves the two clusters is determined by a cost-sensitive version of linear SVM:
\begin{equation}
\label{eq9}
\underset{\bm{w}}{\min}\left\|\bm{w}\right\|_2+C\sum_{k=1}^2 f(p_k)\sum_{\bm{x}_i\in\{\mathcal{S}_j^k\}} \left( max(0,1-z_i\bm{w}^T \bm{x}_i)\right)^2,
\end{equation}
where $p_k=|\mathcal{S}_j^k|/|\mathcal{S}_j|$ denotes the cluster proportion, $\bm{w}$ is the weight vector, $C$ is a regularization parameter, and $z_i=1$ if $\bm{x}_i\in\mathcal{S}_j^1$ and -1 otherwise. Each sample is finally sent to either $\mathcal{S}_j^l$ or $\mathcal{S}_j^r$ by $sgn(\bm{w}^T \bm{x}_i)$. The resulting splitting function is thus {\em learned} in a cost-sensitive manner instead of being chosen from some predefined splitting rules. Note the cost here is defined as a function of the cluster distribution rather than the targeted class distribution, but they will correlate well at the deeper tree depth with much purer nodes where Eq.~\ref{eq9} can better play its role.

In regression trees, we perform a cost-sensitive regression at each node $\mathcal{S}_j$ using a weighted linear SVR:
\begin{equation}
\label{eq10}
\underset{\bm{w}}{\min}\left\|\bm{w}\right\|_2+C\sum_{y\in\mathcal{C}} f(p_y)\sum_{y_i=y \atop \bm{x}_i\in\mathcal{S}_j} \left( max(0,|y_i-\bm{w}^T \bm{x}_i|-\varepsilon)\right)^2,
\end{equation}
where $\varepsilon\geq0$, and we directly penalize the true label distribution $\{p_y=|\{y_i=y,\bm{x}_i\in\mathcal{S}_j\}|/|\mathcal{S}_j|\}$ as costs. The node then branches left if the numeric prediction $\{\bm{w}^T \bm{x}_i\}$ is smaller than the local mean of labels $\sum_{\bm{x}_i\in\mathcal{S}_j}y_i/|\mathcal{S}_j|$, otherwise branches right.

In practice, we use the cost transformation technique in~\cite{Chang11} to solve the above weighted SVM/SVR. The cost function $f(\cdot)$ is defined by a reciprocal decreasing function as $f(p)=(1-p)/p$. Obviously, $f(p)$ gives larger weights to the minority classes which proves effective to improve their prediction accuracies without losing the overall performance in our experiments. In addition, we use the inverse class frequencies to reweight the information gain (Eq.~\ref{eq8}, as in~\cite{Chen04}) to select the best $D'$ random features in both classification and regression trees. The result is a CS-RF framework able to carve reasonably good local neighborhoods for both the majority and minority classes.

\begin{algorithm}[t]
\caption{\textbf{: CS-RF-Induced DSNA}}
\label{alg1}
\begin{algorithmic}
\renewcommand{\algorithmicrequire}{\textbf{Input:}}
\REQUIRE Training set $\left\{(\bm{x}_i,y_i)\right\}_{i=1}^N$, trained CS-RF, query $\bm{q}$.\\
\renewcommand{\algorithmicensure}{\textbf{Initialization:}}
\ENSURE to predict $y$ of $\bm{q}$\\
$\bullet$ Merge for $\bm{q}$ all its reached leaf samples to $\mathcal{R}$.\\
$\bullet$ Via \textbf{Discriminative Local Clustering}, obtain clusters $\{\mathcal{L}_k\}_{k=1}^K$, features $\{\bm{L}_k\}_{k=1}^K$ and labels $\{\bm{y}_k\}_{k=1}^K$. \\
$\bullet$ Set $y^{(0)}$ as the mean of $\bm{y}_k$ for regression or its majority vote for classification in $\mathcal{AH}_k$ that best approximates $\bm{q}$. \\
\renewcommand{\algorithmicensure}{\textbf{Outer Loop:}}
\ENSURE Iterate on $t=1,\ldots,T$ until convergence\\
$\bullet$ Update $\{k^{(t-1)},\bm{v}_k^{(t-1)}\}$ by Eq.~\ref{eq2} as the ones that best approximate $\bm{q}$.\\
$\bullet$ Update the sparse coefficient estimate $\overline{\bm{\alpha}}_k^{(t-1)}$ by Eq.~\ref{eq4}.\\
$\bullet$ Update $\bm{\alpha}_k^{(t-1)}$ via \textbf{Sparse Neighbor Approximation} by minimizing Eq.~\ref{eq6}. \\
$\bullet$ Predict label $y^{(t)}=\bm{y}_k^T\bm{\alpha}_k^{(t-1)}$ or by majority voting among $\bm{y}_k$ with nonzero coefficients.\\
\renewcommand{\algorithmicensure}{\textbf{Output:}}
\ENSURE Converged label $\hat{y}$.\\
\end{algorithmic}
\end{algorithm}

\subsection{Convergence and Complexity}

\begin{figure}
\begin{center}
\includegraphics[width=0.8 \linewidth]{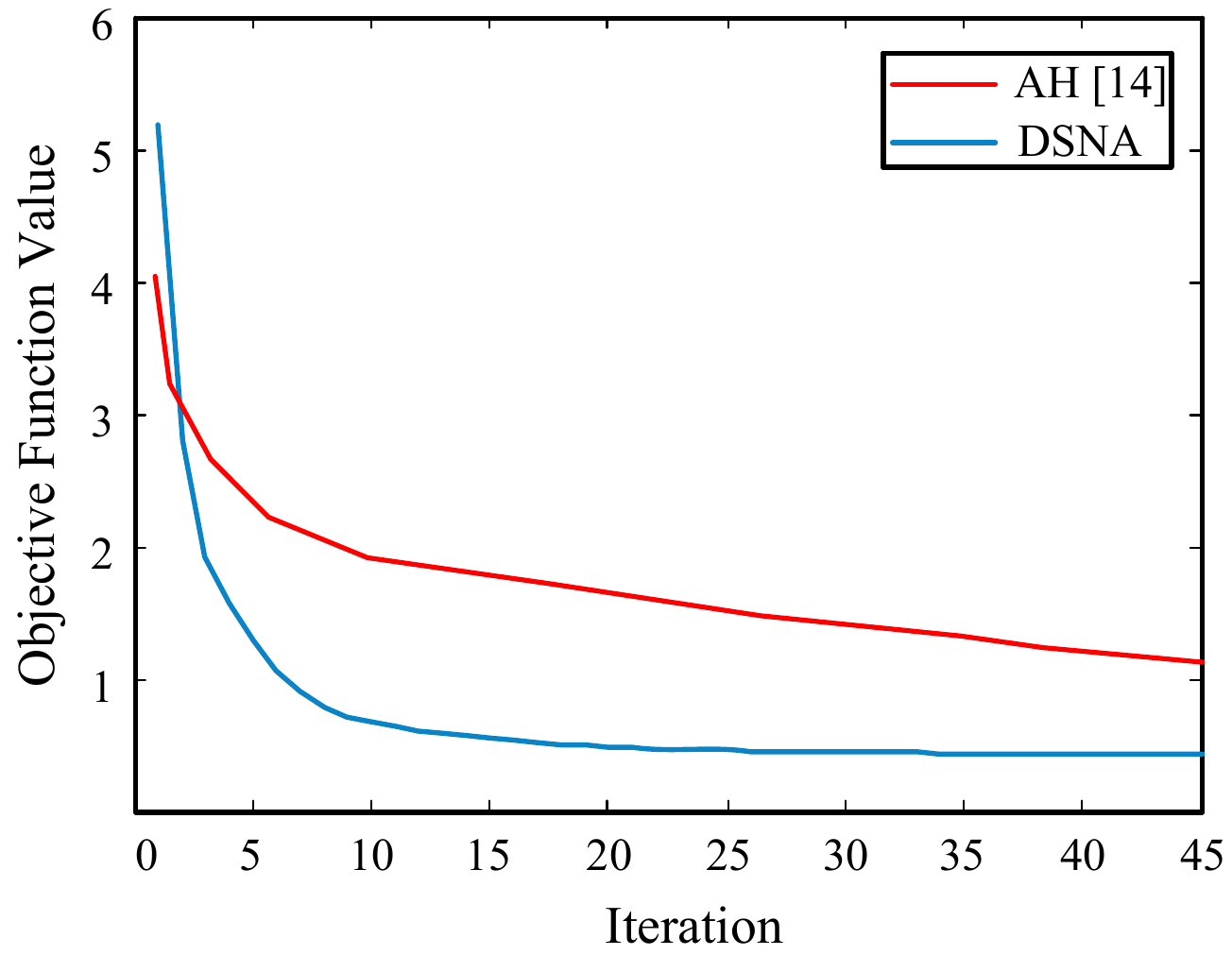}
\end{center}
\vskip -0.5cm
\caption{Comparison of the convergence of unsupervised AH~\cite{Hu11} and our DSNA given a query in the age estimation task.}
\label{fig5}
\end{figure}

Our full algorithm is detailed in Algorithm \ref{alg1}. Similar to the affine hull (AH) method~\cite{Hu11}, Algorithm \ref{alg1} can converge to a global solution. Compared with AH's non-asymptotic convergence rate of $O(1/t^2)$, our DSNA method converges even faster as shown in Figure~\ref{fig5}. Typically DSNA converges within 10 iterations with lower objective values thanks to the introduced class discrimination as a guidance. In our experiments, we will visualize some converged examples with accurate predictions in different vision tasks. As there are usually dozens or hundreds of retrieved samples in the initial data neighborhood, DSNA typically takes only 0.5s to run on an Intel Core i7 4.0GHz CPU.

\section{Experiments}
\label{sec4}
We validate the effectiveness of our CS-RF-induced DSNA method in three vision tasks at various imbalance levels: the high-level tasks of age estimation and head pose estimation (by regression) and the low-level task of edge detection (by classification).
\subsection{Experimental Settings}
\noindent
{\bf Dataset settings:} For age estimation, the FG-NET~\cite{Geng07} and MORPH~\cite{Chang11} datasets are used. FG-NET contains 1002 facial images of 82 subjects with ages in a range from 0 to 69. Algorithms are evaluated by the leave-one-person-out protocol. MORPH contains about 55000 images of more than 13000 subjects with ages between 16 and 77. We randomly split it into three disjoint subsets S1, S2 and S3 as in~\cite{Dong14}. Algorithms repeat 1) training on S1, testing on S2+S3 and 2) training on S2, testing on S1+S3 with the average result reported. Both datasets are highly imbalanced (see Figure~\ref{fig1}(a)) and class-overlapped. FG-NET further suffers from the issue of small data. For both, we use AAM~\cite{Cootes01} as the feature extractor, and Mean Absolute Error (MAE) as the evaluation metric.

For head pose estimation, poses should intrinsically admit an imbalanced distribution with much more near-frontal instances than the profile ones. Unfortunately, we are unable to obtain such datasets with ground truth labels (\eg~``Face Pose'' dataset~\cite{AghajanianP09}) for experiments. We adopt the popular Pointing'04 dataset instead that exhibits some imbalance in pitch angles. The dataset contains images from 15 subjects each with two series of 93 pose images. The pose is discretized into 9 pitch angles $\{ \pm90^{\circ},\pm60^{\circ},\pm30^{\circ},\pm15^{\circ},0^{\circ} \}$ and 13 yaw angles $\{ \pm90^{\circ},\pm75^{\circ},\pm60^{\circ},\pm45^{\circ},\pm30^{\circ},\pm15^{\circ},0^{\circ} \}$. However, when the pitch angles are $\{ \pm90^{\circ}\}$, the yaw is always $\{0^{\circ} \}$ (so $7\times13+2=93$ poses in total), leading to an imbalance ratio of 1:13 between $\{ \pm90^{\circ}\}$ pitch angles and others. We further test in the case where pitch angles are randomly removed to form a Gaussian-like distribution to mimic the real-world imbalanced distribution. As in~\cite{Haj12,Hara14}, evaluation of MAE is performed with 5-fold cross-validation using HOG features.

\begin{figure}[!t]
\begin{center}
\includegraphics[width=1.0 \linewidth]{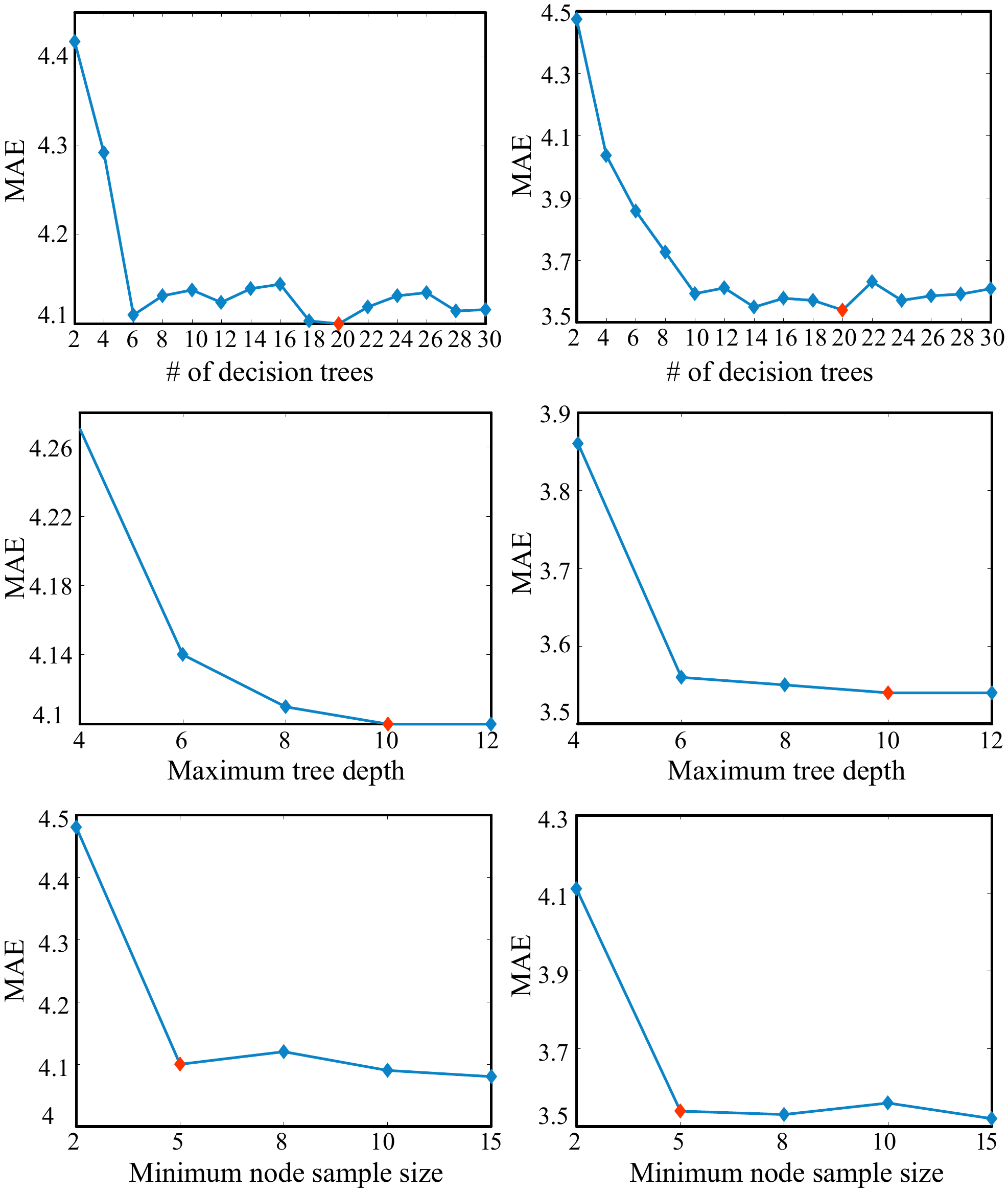}
\end{center}
\vskip -0.5cm
\caption{Parameter sweeps for age estimation (left column) and head pose estimation (right column). Each row (from top to bottom) considers the parameter of tree number, maximum tree depth and minimum node sample size, respectively. The chosen parameter value is marked in red.}
\label{fig6}
\end{figure}

For edge detection, we use the BSDS500~\cite{Pablo11} and NYUD (v2)~\cite{Silberman12} datasets, the latter for testing cross-dataset generalization. BSDS500 contains 200 training, 100 validation and 200 testing images. NYUD contains 1449 pairs of RGB and depth images. We follow~\cite{ren12} to use 60\%/40\% training/testing split (1/3 training data for validation) with the images reduced to $320\times240$ pixels. For cross-dataset testing, we only use RGB images on both datasets. We combine our method in classification mode with the structured edge detector~\cite{Dollar15} since it induces classification forest like us but operates on edge patches instead of pixels, which proves efficient in practice. We use the same multiple low-level features extracted from $32\times32$ image patches and apply non-maximal suppression prior to evaluation as in~\cite{Dollar15}. Edge detection accuracy is evaluated by: fixed contour threshold (ODS), per-image best threshold (OIS), and average precision (AP)~\cite{Pablo11}.

\begin{figure}[!t]
\begin{center}
\includegraphics[width=1.0 \linewidth]{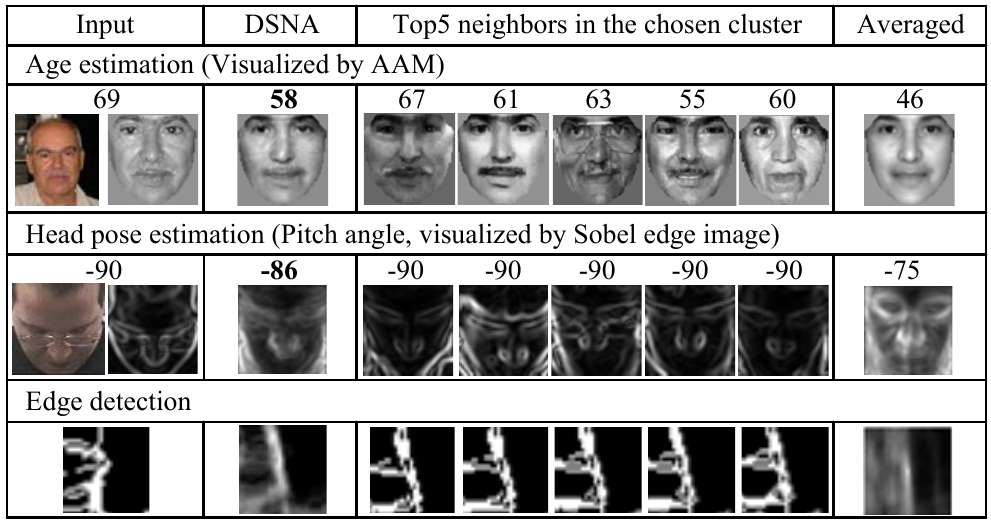}
\end{center}
\vskip -0.5cm
\caption{Visualizations of both the DSNA converged result and simple averaged result on the retrieved samples by CS-RF. Results are shown for the minority class testing samples in all the three tasks.}
\label{fig7}
\end{figure}

\noindent
{\bf Parameters:} For age and head pose estimation, we empirically combine 20 cost-sensitive trees in our regression forest, and terminate splitting when the maximum depth 10 is reached or if the node sample size is fewer than 5. Figure~\ref{fig6} shows the robustness of these parameters across tasks. Evaluations are done by varying one parameter at a time, with others fixed. The chosen parameter value is marked in red. For edge detection, our method is combined with~\cite{Dollar15} and uses the same parameter settings.

Cross-validation is used to determine the trade-off parameter $C$ for cost-sensitive SVM/SVR (Eq.~\ref{eq9} and \ref{eq10}), $\tau$ for biased distance (Eq.~\ref{eq1}), $\lambda$ and $\gamma$ in Eq.~\ref{eq5}. We select $K$ for discriminative local clustering from 2 to 4.

\begin{table}[!t]
\addtolength{\tabcolsep}{-4pt}
\caption{Ablation Test and Comparisons of CS-RF and DSNA in Age Estimation (MAE on FG-NET), Head Pose Estimation (Avg. MAE on Pointing'04) and Edge Detection (ODS on BSDS500, Higher is Better).}
\begin{center}
\resizebox{1.0\linewidth}{!}{
\begin{tabular}
{ c !{\vrule width1pt} c c !{\vrule width1pt} c c c !{\vrule width1pt} c c }
\Xhline{1pt}
 & & RF+ & RED- & WRF &  & CS-RF+ & CS-RF+ \\
\raisebox{1ex}[0pt]{Methods} & \raisebox{1ex}[0pt]{RF} & SMOTE~\cite{Chawla02} & SVM~\cite{LiL06} & \cite{Chen04} & \raisebox{1ex}[0pt]{CS-RF} & AH~\cite{Hu11} & DSNA \\ \hline \Xhline{1pt}
Age & 5.28 & 5.39 & 5.24 & -- & \textbf{4.81} & 4.89 & \textbf{4.10} \\
Pose & 6.41 & 6.65 & 6.53 & -- & \textbf{4.02} & 4.28 & \textbf{3.54} \\
Edge & 0.75 & 0.75 & -- & 0.75 & \textbf{0.76} & 0.76 & \textbf{0.78} \\ \Xhline{1pt}
\end{tabular}
}
\end{center}
\label{tb1}
\end{table}

\begin{table*}[!t]
\addtolength{\tabcolsep}{-3pt}
\caption{Comparisons of Age Estimation Results (MAE) on FG-NET and MORPH Datasets.}
\begin{center}
\resizebox{\linewidth}{!}{
\begin{tabular}{ c c c c c c c !{\vrule width1pt} c c }
\Xhline{1pt}
\multicolumn{7}{c!{\vrule width1pt}}{FG-NET} & \multicolumn{2}{c}{MORPH} \\ \hline \Xhline{1pt}
RUN~\cite{Yan07} & RED-SVM~\cite{LiL06} & MTWGP~\cite{Zhang10} & BIF~\cite{Guo09} & CPNN~\cite{Geng13} & CSOHR~\cite{Chang15} & CA-SVR~\cite{Ke13} & KPLS~\cite{Guo11} & KCCA~\cite{Guo13}\\
5.33 & 5.24 & 4.83 & 4.77 & 4.76 & 4.70 & 4.67 & 4.04 & 3.98 \\ \Xhline{1pt}
Choi~\etal~\cite{Choi11} & MidFea-NS~\cite{KongJY14} & Han~\etal~\cite{Han13} & RealAdaBoost~\cite{Ren14} & OHRank~\cite{Chang11} & lsRCA~\cite{Chao13} & CS-RF+DSNA & MSCNN~\cite{Dong14} & CS-RF+DSNA \\
4.66 & 4.62 & 4.60 & 4.49 & 4.48 & 4.38 & \textbf{4.10} & 3.63 & \textbf{3.54} \\ \Xhline{1pt}
\end{tabular}
}
\end{center}
\label{tb2}
\end{table*}

\subsection{Evaluation of the CS-RF and DSNA}
We start with evaluating our key components of CS-RF and DSNA. CS-RF concerns about generating good local neighborhood, while DSNA makes unbiased and extrapolative prediction and is the major contribution of this paper.

Figure~\ref{fig7} visualizes the benefit of DSNA over simple averaged prediction in the three considered tasks. Clearly, given an appropriate local neighborhood, \eg~by CS-RF, DSNA can localize the correct mode (cluster) in it for the difficult minority class samples, thus making much more unbiased predictions than by simply averaging. More significantly, for age estimation on the small FG-NET dataset, there are very few elderly samples with many missing classes, but our DSNA extrapolates well from the limited data.

Table~\ref{tb1} quantifies the benefits of both CS-RF and DSNA against other competitive schemes in the three tasks. Note all the RF variants in the left and middle columns---RF+SMOTE, WRF (Weighted RF) and CS-RF simply average tree predictions as in standard RF. They do not consider data extrapolation as AH (Affine Hull)~\cite{Hu11} and DSNA do. We make the following observations: 1) The over-sampling method SMOTE shows no benefits over Bagging in standard RF since it can introduce undesirable noise (\eg~in age and pose cases). 2) Cost-sensitive learning, in the middle column, helps for these imbalanced tasks, and our CS-RF consistently outperforms RED-SVM and WRF. This suggests that simple weighting schemes in RED-SVM and WRF are not adequate in complex imbalanced tasks. In contrast, our CS-RF can be seen as an ensemble of cost-sensitive experts organized in hierarchical trees, with higher capability and robustness. Another advantage is that CS-RF provides a unified cost-embedded solution to both regression and classification. 3) The supervised DSNA combined with CS-RF leads to large improvements, whereas the unsupervised AH shows no improvements or even worse results. This emphasizes the importance of using supervisory information. DSNA uses such information intelligently by extrapolating from several discriminatively trained AH models with a class-aware constraint (Eq.~\ref{eq5}).

\subsection{Comparison with State-of-the-Arts}
\noindent
{\bf Age estimation:}
We compare with the state-of-the-arts on FG-NET and MORPH datasets in Table~\ref{tb2}. Our CS-RF+DSNA outperforms most methods by a large margin, and reduces the MAEs of the runner-up lsRCA and MSCNN on the two datasets by 6.4\% and 2.5\% respectively. The larger improvement on FG-NET is impressive because the dataset is very small and has missing class labels (old ages). This validates our competence in synthesizing novel labels on small imbalanced datasets. Note the mere cost-sensitive methods RED-SVM, CSOHR and OHRank all show their inferiority on this imbalanced dataset, necessitating the ability of extrapolation. The advanced features---Bio-Inspired Features (BIF), generalized BIF with scattering transform in~\cite{Chang15} and feature selection by RealAdaBoost~\cite{Ren14} also do not reach top in this task. In contrast, our method, using the AAM features only, even outperforms the deep feature-based MidFea-NS and MSCNN due to the handling of data imbalance. Compared with the indirect imbalance-handling methods, CPNN, CA-SVR and lsRCA, ours performs much better by introducing explicit mechanisms that are discriminative and extrapolative.

\begin{figure}[!t]
\begin{center}
\includegraphics[width=0.75 \linewidth]{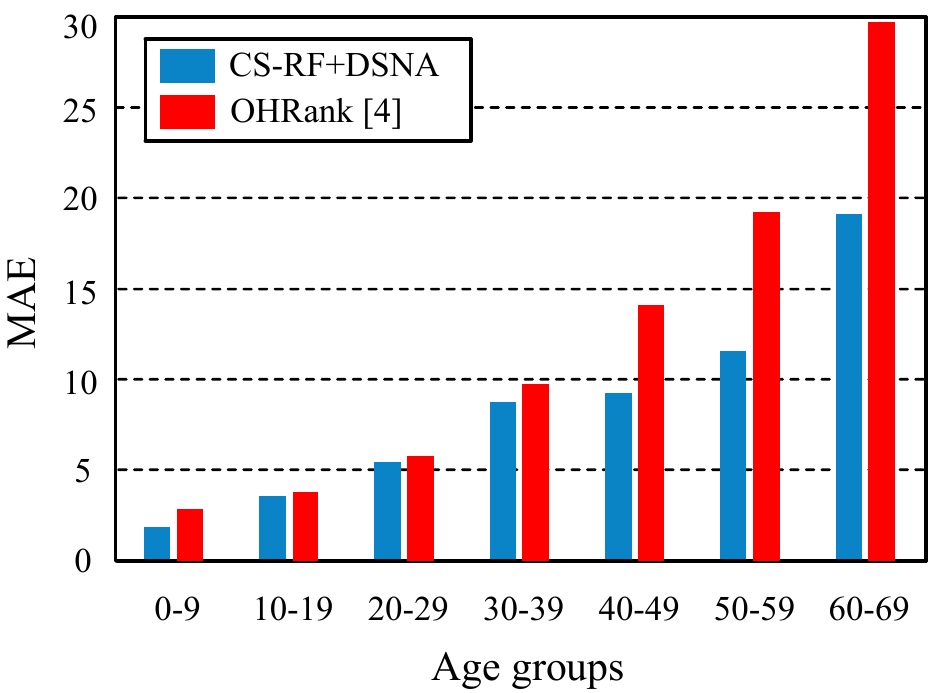}
\end{center}
\caption{MAEs at different age groups on the FG-NET dataset.}
\label{fig8}
\end{figure}

Figure~\ref{fig8} shows the MAEs per decade of CS-RF+DSNA and the competitive OHRank with public implementation. From the comparison, the benefits of our method become prominent at old ages with extremely limited samples. We attribute the performance gain to the extrapolative ability of the proposed DSNA.

\begin{table}[!t]
\caption{Comparison of Pose Estimation MAEs[$^{\circ}$] on Pointing'04.}
\begin{center}
\resizebox{0.65 \linewidth}{!}{
\begin{tabular}{ c !{\vrule width1pt} c c c }
\Xhline{1pt}
Method & Yaw & Pitch & Avg. \\
\Xhline{1pt}
KPLS~\cite{Haj12} & 6.56 & 6.61 & 6.59 \\
SLDML~\cite{Liu14} & 6.31 & 6.71 & 6.51 \\
Fenzi~\etal~\cite{Fenzi13} & 5.94 & 6.73 & 6.34 \\
GLLiM~\cite{drouard15} & 5.62 & 6.68 & 6.15 \\
KRF~\cite{Hara14} & 5.29 & 2.51 & 3.90 \\
CS-RF+DSNA & \textbf{5.04} & \textbf{2.03} & \textbf{3.54} \\
\Xhline{1pt}
\end{tabular}
}
\end{center}
\label{tb3}
\end{table}

\begin{figure}[!t]
\begin{center}
\includegraphics[width=1.0 \linewidth]{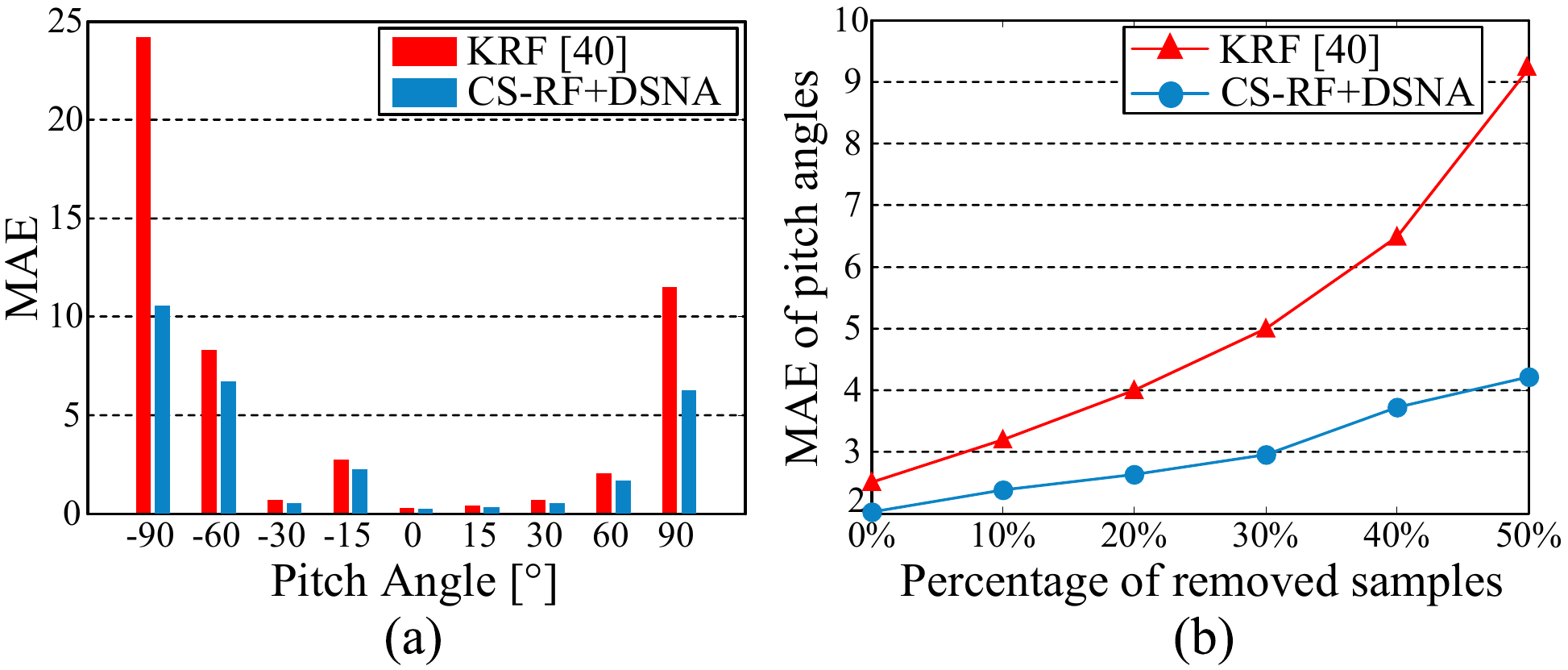}
\end{center}
\caption{Comparisons of imbalanced pitch angle estimation on Pointing'04. (a) MAEs at different pitch angles. (b) Average pitch MAEs with different percentages of samples removed.}
\label{fig9}
\end{figure}

\noindent
{\bf Head pose estimation:}
Table~\ref{tb3} compares our method with the regression-based prior arts KPLS, SLDML, Fenzi~\etal, GLLiM and KRF on Pointing'04 dataset. As mentioned in Section~\ref{sec2}, the sparse sampling of pose angle compounds the learning difficulty, especially for the imbalanced pitch angles. Our method performs best again for both pose angles, with a large margin for pitch. Figure~\ref{fig9}(a) further compares our results at different pitch angles with those of KRF, the state-of-the-art regression forest-based method. Our method benefits from the proposed cost-sensitive (CS) RF and extrapolative DSNA, thus better handling the data imbalance at the rare $\pm90^{\circ}$ poses. Specifically, we obtain a MAE of 8 degrees for those $\pm90^{\circ}$ poses with only 24 training samples (hundreds of samples for other poses), which is 55.6\% lower than that of KRF. The larger MAE at $-90^{\circ}$ is due to the higher variations of this angle compared to that of $90^{\circ}$. We finally compare with KRF in Figure~\ref{fig9}(b) where pitch samples are randomly removed to form a Gaussian-like distribution aiming to mimic the real-world one. It can be observed that the performance of our method degrades more gracefully with the increase of removed data, showing a strong ability to handle small imbalanced data.

\begin{table}[!t]
\caption{Comparison of Edge Detection Results on the BSDS500 Dataset.}
\begin{center}
\resizebox{0.7 \linewidth}{!}{
\begin{tabular}{  c !{\vrule width1pt} c c c  }
\Xhline{1pt}
Method & ODS & OIS & AP \\
\Xhline{1pt}
ISCRA~\cite{ren13} & 0.72 & 0.75 & 0.46 \\
gPb-owt-ucm~\cite{Pablo11} & 0.73 & 0.76 & 0.73 \\
Sketch Tokens~\cite{LimCVPR13} & 0.73 & 0.75 & 0.78 \\
SCG~\cite{ren12} & 0.74 & 0.76 & 0.77 \\
PMI+sPb~\cite{Isola14} & 0.74 & 0.77 & 0.78 \\
SE~\cite{Dollar15} & 0.75 & 0.77 & 0.80 \\
OEF~\cite{HallmanF15} & 0.75 & 0.77 & 0.82 \\
SE+multi-ucm~\cite{APBMM2014} & 0.75 & 0.78 & 0.76 \\ \hline
DeepNet~\cite{KivinenWH14} & 0.74 & 0.76 & 0.76 \\
N$^4$-Fields~\cite{Ganin14} & 0.75 & 0.77 & 0.78 \\
DeepEdge~\cite{gberta15_CVPR} & 0.75 & 0.77 & 0.81 \\
DeepContour~\cite{Shen15} & 0.76 & 0.78 & 0.80 \\
HFL~\cite{gberta15_ICCV} & 0.77 & 0.79 & 0.80 \\
HED~\cite{xie15} & \textbf{0.78} & \textbf{0.80} & \textbf{0.83} \\\hline
CS-SE+DSNA & 0.77 & 0.79 & 0.81 \\
\Xhline{1pt}
\end{tabular}
}
\end{center}
\label{tb4}
\end{table}

\noindent
{\bf Edge detection:}
In this task, severe imbalance exists between the positive edge classes and negative non-edge class. We refer our combined method with structured edge (SE) detector~\cite{Dollar15} as CS-SE+DSNA.

Table~\ref{tb4} summarizes extensive comparisons with state-of-the-art methods on BSDS500. It is observed that CS-SE+DSNA outperforms all ``shallow'' methods (top cell) across all evaluation metrics, and also performs better than most deep models (bottom cell) and is comparable to the top HED. The results are impressive since our method only uses hand-designed features.
In is worth mentioning that HED utilizes a deep CNN with as many as 16 layers for good performance. However our ``shallow'' method can still compete with such deep models due to the proposed mechanism for handling imbalance.

\begin{figure*}[!t]
\begin{center}
\includegraphics[width=1.0 \linewidth]{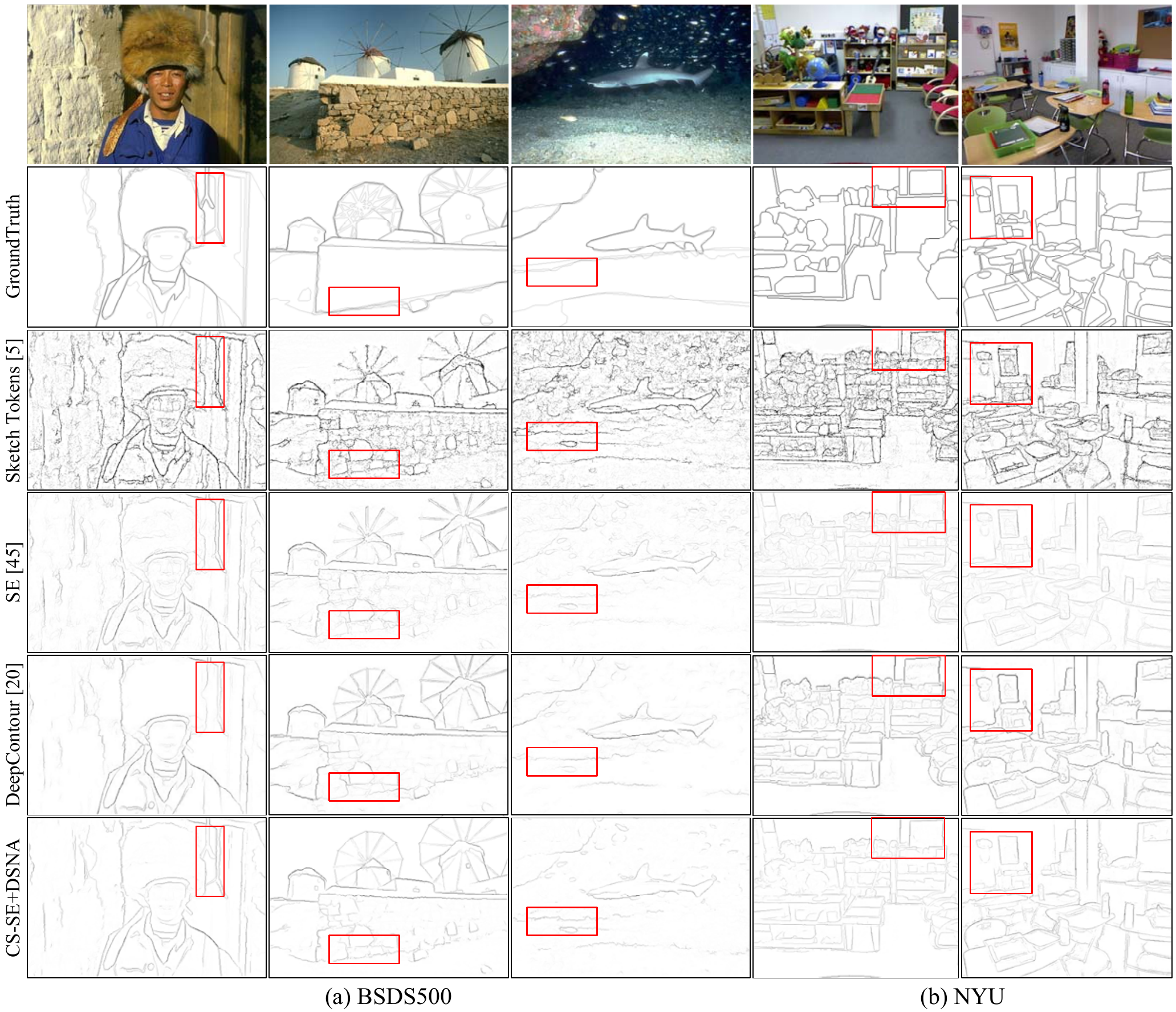}
\end{center}
\vskip -0.25cm
\caption{Edge detection results on the (a) BSDS500 dataset and (b) NYU dataset with BSDS trained model.}
\label{fig10}
\end{figure*}

This advantage of our method also holds with respect to the compared non-deep methods. Among them, Sketch Tokens, SE and OEF are based on random forest similar to ours. The performance gain compared to their results can thus be attributed to the capability of correctly classifying imbalanced edge patches and generalizing to novel edge structures. Figure~\ref{fig10}(a) confirms this standpoint by comparing CS-SE+DSNA with three related random forest-based methods, including DeepContour that applies random forest on top of deeply learned features. Clearly, CS-SE+DSNA is able to produce cleaner results with preserved edge structures. In other words, it is capable of predicting the minority edges without jeopardizing the majority non-edges that make edge maps clean. Also, the computational overhead is modest as compared to SE.

To further validate the extrapolative ability of our method, we perform cross-dataset generalization tests in comparison to the competing methods with public results. The NYU/NYU results are used as baselines, see Table~\ref{tb5}. In both cases of NYU/NYU and BSDS/NYU testing, we find favorable performance, demonstrating a superior capability of generalization. Figure~\ref{fig10}(b) shows the visual results.

\begin{table}[!t]
\caption{Comparison of Edge Detection (Top) and Cross-Dataset Generalization (Bottom) Results on the NYU Dataset Using only RGB Images. TRAIN/TEST Indicates the Training/Tesing Dataset Used.}
\begin{center}
\resizebox{0.85 \linewidth}{!}{
\begin{tabular}{ c !{\vrule width1pt} c c c }
\Xhline{1pt}
Method & ODS & OIS & AP \\
\Xhline{1pt}
gPb~\cite{Pablo11} (NYU/NYU) & 0.51 & 0.52 & 0.37 \\
SCG~\cite{ren12} (NYU/NYU) & 0.55 & 0.57 & 0.46 \\
SE~\cite{Dollar15} (NYU/NYU) & 0.60 & 0.61 & 0.56 \\
CS-SE+DSNA (NYU/NYU) & \textbf{0.62} & \textbf{0.63} & \textbf{0.60} \\ \Xhline{1pt}
SE~\cite{Dollar15} (BSDS/NYU) & 0.55 & 0.57 & 0.46 \\
DeepContour~\cite{Shen15} (BSDS/NYU) & 0.55 & 0.57 & 0.49 \\
CS-SE+DSNA (BSDS/NYU) & \textbf{0.57} & \textbf{0.58} & \textbf{0.51} \\
\Xhline{1pt}
\end{tabular}
}
\end{center}
\label{tb5}
\end{table}

\section{Conclusion}
\label{sec5}

We propose in this paper a principled method for handling data imbalance by discriminative sparse neighbor approximation. With this method, we are able to make unbiased predictions with preserved discriminative and extrapolative ability. Such predictions are made among the local data neighborhood retrieved by a cost-sensitive decision forest. Our method proves effective in diverse vision tasks at various imbalance levels, and substantially outperforms the state-of-the-arts including some deep learning methods that ignore the imbalance issue. Our method shows its great potential as an efficient and general purpose solution for imbalanced learning. Future works include pushing the framework deeper by using cascaded forests with multi-level predictions, to explore the extent to which we can achieve by simulating deep architectures.

\bibliographystyle{IEEEtran}
\bibliography{egbib}

%

\begin{IEEEbiography}[{\includegraphics[width=1in,height=1.25in,clip,keepaspectratio]{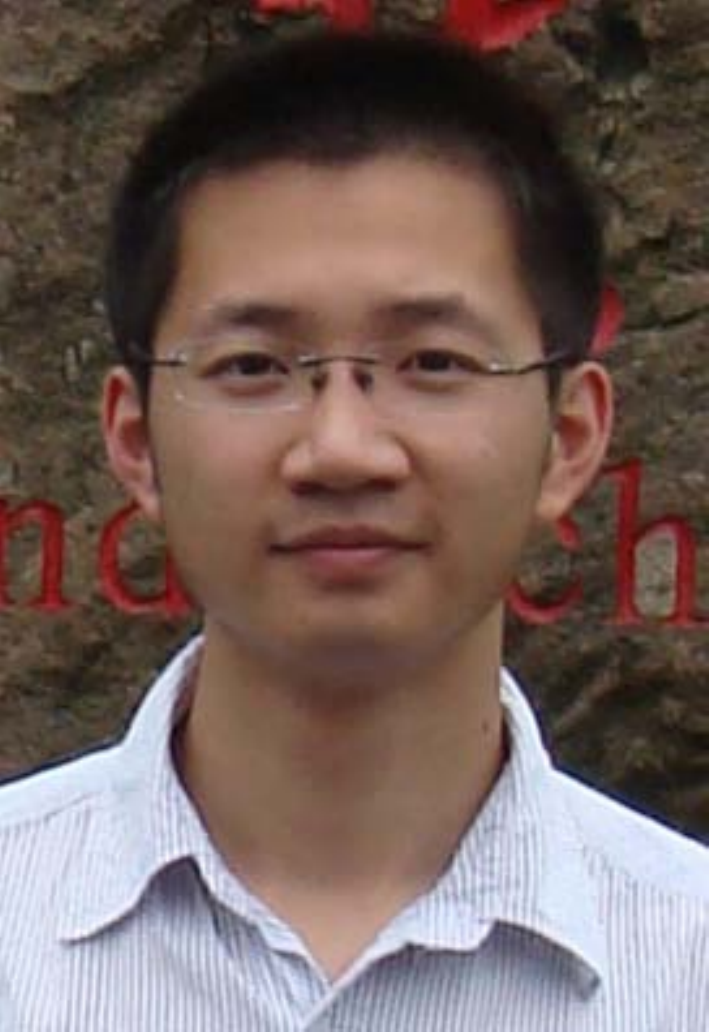}}]{Chen Huang}
received the Ph.D. degree in Electronic Engineering from Tsinghua University, Beijing, China, in 2014. He is currently a postdoctoral research associate in the Department of Information Engineering of the Chinese University of Hong Kong. His research interests include computer vision, pattern recognition and image processing, with focus on deep learning, face analysis and recognition.
\end{IEEEbiography}


\begin{IEEEbiography}[{\includegraphics[width=1in,height=1.25in,clip,keepaspectratio]{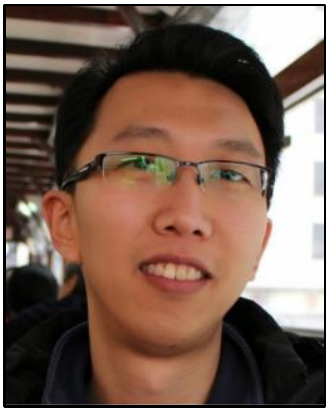}}]{Chen Change Loy}
received the PhD degree in Computer Science from the Queen Mary University of London in 2010. He is currently a Research Assistant Professor in the Department of Information Engineering, Chinese University of Hong Kong. Previously he was a postdoctoral researcher at Vision Semantics Ltd from 2011 2013. His research interests include computer vision and pattern recognition, with focus on face analysis, deep learning, and visual surveillance.
\end{IEEEbiography}

\begin{IEEEbiography}[{\includegraphics[width=1in,height=1.25in,clip,keepaspectratio]{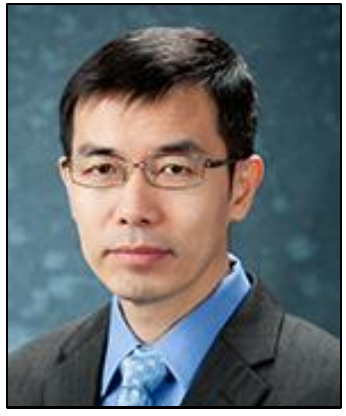}}]{Xiaoou Tang}
received the B.S. degree from the University of Science and Technology of China, Hefei, in 1990, and the  M.S. degree from the University of Rochester, Rochester, NY, in  1991. He received the Ph.D. degree from the Massachusetts Institute of Technology, Cambridge, in 1996. He is a Professor and the Chairman of the Department of Information Engineering. He worked as the group manager of the Visual Computing Group at the Microsoft Research Asia from 2005 to 2008. His research interests include computer vision, pattern recognition, and video processing. Dr. Tang received the Best Paper Award at the IEEE Conference on Computer Vision and Pattern Recognition (CVPR) 2009. He is a program chair of the IEEE International Conference on Computer Vision (ICCV) 2009 and has served as an  Associate Editor of IEEE Transactions on Pattern Analysis and Machine Intelligence (PAMI) and International Journal of Computer Vision (IJCV). He is a Fellow of IEEE.
\end{IEEEbiography}


\vfill


\end{document}